\RequirePackage{fix-cm}
\documentclass[natbib,smallextended]{svjour3}       
\smartqed  
\usepackage{graphicx}
\usepackage{tikz-qtree}
\usepackage{url}
\usepackage{array}
\usepackage{enumerate}

\newcolumntype{P}[1]{>{\centering\arraybackslash}p{#1}}

\journalname{}

\newcommand{\GS}[1]{\textbf{GS: #1 }}
\begin{document}

\title{CLAUDETTE: an Automated Detector of Potentially Unfair Clauses in Online Terms of Service
}

\titlerunning{CLAUDETTE}        

\author{Marco Lippi \and Przemyslaw Palka \and Giuseppe Contissa \and Francesca Lagioia \and Hans-Wolfgang Micklitz \and Giovanni Sartor \and Paolo Torroni}

\authorrunning{Author list suppressed} 

\institute{M. Lippi \at
              DISMI -- Universit\`a di Modena e Reggio Emilia, Italy \\
              Tel.: +390522522660\\
              \email{marco.lippi@unimore.it}           
           \and
           P. Palka, F. Lagioia, H.-W. Micklitz \at
              Law Department, European University Institute, Florence, Italy \\
              \email{\{przemyslaw.palka, francesca.lagioia, hans.micklitz\}@eui.eu}           
	   \and
	   G. Contissa, G. Sartor \at
	      CIRSFID, Alma Mater -- Universit\`a di Bologna, Italy \\
              \email{\{giuseppe.contissa, giovanni.sartor\}@unibo.it}           
	   \and
	   P. Torroni \at
	      DISI, Alma Mater -- Universit\`a di Bologna, Italy \\
              \email{paolo.torroni@unibo.it}           
\and
This is a pre-print of an article published in Artificial Intelligence and Law. The final authenticated version is available online at: https://doi.org/10.1007/s10506-019-09243-2.
}

\date{Received: date / Accepted: date}

\maketitle

\begin{abstract}
Terms of service of on-line platforms too often contain clauses that are potentially unfair to the consumer. We present an experimental study where machine learning is employed to automatically detect such potentially unfair clauses. Results show that the proposed system could provide a valuable tool for lawyers and consumers alike.
\end{abstract}

\section{Introduction} \label{sec:introduction}

A recent survey on policy-reading behaviour~\citep{obar2016biggest} reveals that consumers rarely read the contracts they are required to accept. This resonates with our direct experience and with what has long been said, that the biggest lie on the Internet is ``I have read and agree to the terms and conditions.'' We use smartphones to gather and share information, connect on social media, entertain ourselves, check our online banking and so on. Virtually every app we install and website we browse have their own Terms of Service (ToS). These are contracts that bind us by the time we switch on the phone or browse a website, although we are not necessarily aware of what we just agreed upon.

There are reasons why many consumers do not read or understand ToS, as well as privacy policies or end-user license agreements (EULA)~\citep{bakos2014does}. Reports indicate that such documents can be overwhelming to the few consumers who actually venture to read them~\citep{DoC2010}. It has been estimated that actually reading the privacy policies alone would carry costs in time of over 200 hours a year per Internet user~\citep{McDonaldCranor}. Another problem is that even if consumers did read the ToS thoroughly, they would have no means to influence their content: the choice is to either agree to the terms offered by a web app or simply not use the service at all.

All this created a need for limitations on traders' contractual freedom, not only to protect consumer interests, but also to enhance the consumers' trust in transnational transactions and improve the common market~\citep{nebbia2007unfair}. European consumer law aims to prevent businesses from using so-called `unfair contractual terms' in the contracts they unilaterally draft and require consumers to accept~\citep{reich2014european}. Law regarding such terms applies also to the ToS of on-line platforms~\citep{loos2016wanted}.
Unfortunately, it turns out such platforms' owners do use in their ToS unfair contractual clauses~\citep{micklitz2017empire}, in spite of the European law, and regardless of consumer protection organizations agencies, which have the competence, but not necessarily the resources, to fight against such unlawful practices.

To address this problem, we propose a machine learning-based method and tool for partially automating the detection of potentially unfair clauses. Such a tool could be useful both for consumer protection organizations and agencies, \GS{by}making their work more effective and efficient, and for consumers, by increasing their understanding of what they agree upon.

This paper builds upon and significantly extends results presented by~\cite{Lippi2017JURIX} in a smaller-scale study where a Support Vector Machine (SVM) was trained on a 20-document corpus. With respect to previous work, the contributions of this study are:
\begin{itemize}
 \item a larger corpus consisting of 50 contracts (over 12,000 sentences), so as to train and evaluate the proposed approach on a wider and more heterogeneous data set;
 \item an extensive comparison of several machine learning systems, including some recent deep learning architectures for text categorization, and a structured SVM for collective classification, which takes into account the sequence of sentences within a document;
 \item a more comprehensive task, which is not restricted to the detection of potentially unfair clauses but also encompasses the classification of such clauses into categories;
 \item the description of a web server, named CLAUDETTE, which we have made available to the community, so as to allow users to submit query documents and analyze the behavior of the system.
\end{itemize}

The paper is organized as follows. In Section~\ref{sec:problem} we first define the problem from the legal point of view. Then, in Section~\ref{sec:dataset} we describe the novel corpus and the document annotation procedure. Section~\ref{sec:methodology} explains the machine learning methodology employed in the system, whereas Section~\ref{sec:results} presents experimental results. Section~\ref{sec:server} describes the web server. Section~\ref{sec:related} discusses related work. Section~\ref{sec:conclusion} concludes with a look to future research.

\section{Problem Description} \label{sec:problem}
In this section we briefly introduce the European consumer law on unfair contractual terms (clauses). We explain what  an unfair contractual term is, present the legal mechanisms created to prevent business from employing \GS{unfair terms}, and describe our contribution to these mechanisms.  

According to art. 3 of the Directive 93/13 on Unfair Terms in Consumer Contracts, a contractual term is unfair if: 1) it has not been individually negotiated; and 2) contrary to the requirement of good faith, it causes a significant imbalance in the partiesâ rights and obligations, to the detriment of the consumer. This general definition is \GS{further} specified in the Annex to the Directive, containing âan indicative and non-exhaustive list of the terms which may be regarded as unfairâ, as well in a few dozen judgments of the Court of Justice of the EU~\citep{micklitz2014court}. Examples of unfair clauses encompass taking jurisdiction away from the consumer, limiting liability for damages on health and/or gross negligence, imposing obligatory arbitration in a country different from consumerâs residence etc. 

Loos and Luzak~\citep{loos2016wanted} identified five categories of \textit{potentially unfair} clauses often appearing in the terms of on-line services: 1) establishing jurisdiction for disputes in a country different than consumerâs residence; 2) choice of a foreign law governing the contract; 3) limitation of liability; 4) the provider's right to unilaterally terminate the contract/access to the service; and 5) the provider's right to unilaterally modify the contract/the service. Our research has identified three additional categories: 6) requiring a consumer to undertake arbitration before the court proceedings can commence; 7) the provider retaining the right to unilaterally remove consumer content from the service, including in-app purchases; 8) having a consumer accept the agreement simply by using the service, not only without reading it, but even without having to click on ``I agree/I accept.'' 

	The 93/13 Directive creates two mechanisms to prevent the use of unfair contractual terms: \textit{individual} and \textit{abstract}  control of fairness. The former takes place when a consumer goes to court:  if a court  finds that a clauses is unfair (which it can do on its own motion), it will consider that the clause is not  binding on the consumer (art. 6). However, most consumers do not take their disputes to courts. That is why abstract fairness control has been created. In each EU Member State, consumer protection organizations have the competence to initiate judicial orin administrative proceedings,  to obtain the declaration that   clauses in consumer contracts are  unfair. The  national implementations of  abstract control  differ in various ways. For instance, consumer protection agencies and/or consumer organisations may be involved to a different degree, there may or may not be fines for using unfair contractual terms, etc.~\citep{schulte2008ec}. One thing that all member states have in common is that if a business uses unfair terms in their contracts, in principle there is always a competent party with the authority to challenge such contracts. 

Unfortunately, the legal mechanism for enforcing the prohibition of unfair contract terms have failed to effectively counter this practice so far.  As reported by some literature~\citep{loos2016wanted}, and as our own research indicates~\citep{micklitz2017empire}, unfair contractual terms are, as of today, widely used in ToS of online platforms. 

In our previous research~\citep{micklitz2017empire}, we developed a theoretical model of tasks that human lawyers currently need to carry out, before starting the legal proceedings concerning the abstract control of fairness of clauses. Those include: 1) finding and choosing the documents; 2) mining the documents for potentially unfair clauses; 3) conducting the actual legal assessment of fairness; 4) drafting the case files and beginning the proceedings. 
Our work aims to automate the second step, enabling a senior lawyer to focus only on clauses that are found by a machine learning classifier to be potentially unfair, thus saving significant time and labor. 

We focus on \textit{potentially} unfair clauses for two reasons. First, we may be unsure whether a certain type of clause falls under the abstract  legislative definition of an ``unfair contractual term''. From a legal standpoint, a given clause can be deemed unfair with absolute certainty only if a competent institution, such as a national court having refereed to the European Court of Justice,  has ruled in that sense. That is the case for certain kinds of clauses, such as a  jurisdiction clause indicating a country different from the consumer's residence,  or limitation of liability for gross negligence~\citep{micklitz2017empire}.
In  other cases the unfairness of a clause has to be argued for, showing that it creates an unacceptable imbalance in  the parties' rights and obligations. A consumer protection body might want to take the case to a court in order to authoritatively establish the unfairness of that clause, but a legal argument for that needs to be created, and the clause may eventually turn out to be judged fair. 
As a second point, unfairness may depend not only on a clause's textual content, but also on the context in which the clause is to be applied. For instance,  a mutual right to unilaterally terminate the contract might be fair in some cases, and unfair in others, for example if unilateral termination would entail losing some digital content (purchased apps, email address, etc.) on the side of the consumer.

\section{Corpus Annotation} \label{sec:dataset}

The corpus consists of 50 relevant on-line consumer contracts, i.e. the Terms of Service of on-line platforms.
Such contracts were selected among those offered by some of the major players in terms of number of users, global relevance, and time of establishment of the service.\footnote{In particular, we selected the ToS offered by: 9gag.com, Academia.edu, Airbnb, Amazon, Atlas Solutions, Betterpoints, Booking.com, Crowdtangle, Deliveroo, Dropbox, Duolingo, eBay, Endomondo, Evernote, Facebook, Fitbit, Google, Headspace, Instagram, Linden Lab, LinkedIn, Masquerade, Microsoft, Moves-app, musically, Netflix, Nintendo, Oculus, Onavo, Pokemon GO, Rovio, Skype, Skyscanner, Snapchat, Spotify, Supercell, SyncMe, Tinder, TripAdvisor, TrueCaller, Twitter, Uber, Viber, Vimeo, Vivino, WhatsApp, World of Warcraft, Yahoo, YouTube and Zynga. The annotated corpus can be downloaded from \url{http://155.185.228.137/claudette/ToS.zip}.}
Such contracts are usually quite detailed in content, are frequently updated to reflect changes both in the service and in the applicable law, and are often available in different versions for different jurisdictions.
in the presence of multiple versions of the same contract, we selected the most recent version available on-line for the European customers.
The mark-up was done in XML. A description of the annotation process follows.

\subsection{Annotation process}

In analyzing the Terms of Service of the selected on-line platforms, we identified eight different categories of unfair clauses. For each type of clause we defined a corresponding XML tag, as shown in Table~\ref{tab:clause_types}.

Notice that not necessarily all the documents contain all clause categories.
For example, Twitter provides two different ToS, the first one for US and non-US residents and the second one for EU residents. The tagged version is the version applicable in the EU and it does not contain any choice of law, arbitration or jurisdiction clauses.

We assumed that each type of clause could be classified   as clearly fair, potentially unfair, or  clearly unfair. In order to mark the different degrees of (un)fairness we appended a numeric value to each XML tag, with 1 meaning clearly fair,  2   potentially unfair, and 3 clarly unfair. Nested tags were used to annotate text segments relevant to more than one type of clause. If one clause covers more then one paragraphs. we chose to tag each paragraph separately, possibly with  different degrees of (un)fairness.

\begin{table}[!bt]
\begin{center}
\caption{Categories of clause unfaireness, with the corresponding symbol used for tagging.\label{tab:clause_types}}
\begin{tabular}{|c|c|c|}
    \hline
    Type of clause & Symbol \\
    \hline
    Arbitration   			& \texttt{<a>} 		\\
    Unilateral change  		& \texttt{<ch>} 	\\
    Content removal  		& \texttt{<cr>} 	\\
    Jurisdiction			& \texttt{<j>} 		\\
    Choice of law 			& \texttt{<law>} 	\\
    Limitation of liability & \texttt{<ltd>} 	\\
    Unilateral termination  & \texttt{<ter>} 	\\
    Contract by using  		& \texttt{<use>} 	\\
    \hline
  \end{tabular}
\end{center}
\end{table}


The \textbf{jurisdiction} clause stipulates what courts will have the competence to adjudicate disputes under the contract.
Jurisdiction clauses giving consumers a right to bring disputes in their place of residence were marked  as clearly fair, whereas clauses stating that any judicial proceeding takes a residence away (i.e. in a different city, different country) were marked as clearly unfair. This assessment is grounded in ECJ's case law, see for example \textit{Oceano} case no. C-240/98.
%
%
An example of jurisdiction clauses is the following one, taken from the Dropbox terms of service:

\begin{quote}
\sloppy \footnotesize
\texttt{<j3> You and Dropbox agree that any judicial proceeding to resolve claims relating to these Terms or the Services will be brought in the federal or state courts of San Francisco County, California, subject to the mandatory arbitration provisions below. Both you and Dropbox consent to venue and personal jurisdiction in such courts.</j3>}
\end{quote}
\begin{quote}
\sloppy \footnotesize
\texttt{<j1>If you reside in a country (for example, European Union member states) with laws that give consumers the right to bring disputes in their local courts, this paragraph doesn't affect those requirements.</j1>}
\end{quote}
The second clause introduces an exception to the general rule stated in the first clause, thus we marked the first one as clearly unfair and the second as clearly fair.

The \textbf{choice of law} clause specifies what law will govern the contract, meaning also what law will be applied in potential adjudication of a dispute arising under the contract.
Clauses defining the applicable law as the law of the consumer's country of residence were marked as clearly fair, as reported in the following examples, taken from the Microsoft services agreements:
\begin{quote}
\sloppy \footnotesize
\texttt{<law1>If you live in (or, if a business, your principal place of business is in) the United States, the laws of the state where you live govern all claims, regardless of conflict of laws principles, except that the Federal Arbitration Act governs all provisions relating to arbitration.</law1>}
\end{quote}
\begin{quote}
\sloppy \footnotesize
\texttt{<law1>If you acquired the application in the United States or Canada, the laws of the state or province where you live (or, if a business, where your principal place of business is located) govern the interpretation of these terms, claims for breach of them, and all other claims (including consumer protection, unfair competition, and tort claims), regardless of conflict of laws principles.</law1>}
\end{quote}
\begin{quote}
\sloppy \footnotesize
\texttt{<law1>Outside the United States and Canada. If you acquired the application in any other country, the laws of that country apply.</law1>}
\end{quote}
In every other case, the choice of law clause was considered as potentially unfair. This is because the evaluation of the choice of law clause needs to take into account several other conditions besides those specified the clause itself (for example, level of protection offered by the chosen law). Consider the following example, taken from the Facebook terms of service:
\begin{quote}
\sloppy \footnotesize
\texttt{<law2>The laws of the State of California will govern this Statement, as well as any claim that might arise between you and us, without regard to conflict of law provisions</law2>}
\end{quote}

The \textbf{limitation of liability}  clause stipulates that the duty to pay damages is limited or excluded, for certain kind of losses, under certain conditions.
Clauses that explicitly affirm non-excludable providers' liabilities were marked as clearly fair. For example, consider the example below, taken from World of Warcraft terms of use:
\begin{quote}
\sloppy \footnotesize
\texttt{<ltd1>Blizzard Entertainment is liable in accordance with statutory law (i) in case of intentional breach, (ii) in case of gross negligence, (iii) for damages arising as result of any injury to life, limb or health or (iv) under any applicable product liability act.</ltd1>}
\end{quote}
Clauses that reduce, limit, or exclude the liability of the service provider were  marked as potentially unfair when concerning  broad categories of losses or causes of them, such as  any harm to the computer system because of malware or loss of data or the suspension, modification, discontinuance or lack of the availability of the service. Also those liability limitation clauses   containing a blanket phrase like  ``to the fullest extent permissible by law'', where considered potentially unfair.  The following example is taken from 9gag terms of service:
%
%
\begin{quote}
\sloppy \footnotesize
\texttt{<ltd2>You agree that neither 9GAG, Inc nor the Site will be liable in any event to you or any other party for any suspension, modification, discontinuance or lack of availability of the Site, the service, your Subscriber Content or other Content.</ltd2>}
\end{quote}
Clause meant to reduce, limit, or exclude the liability of the service provider for physical injuries, intentional damages as well as in case of gross negligence were marked as clearly unfair (based on the Annex to the Directive) as showed in the example below, taken from the Rovio license agreement:
\begin{quote}
\sloppy \footnotesize
\texttt{<ltd3> In no event will Rovio, Rovio's affiliates, Rovio's licensors or channel partners be liable for special, incidental or consequential damages resulting from possession, access, use or malfunction of the Rovio services, including but not limited to, damages to property, loss of goodwill, computer failure or malfunction and, to the extent permitted by law, damages for personal injuries, property damage, lost profits or punitive damages from any causes of action arising out of or related to this EULA or the software, whether arising in tort (including negligence), contract, strict liability or otherwise and whether or not Rovio, Rovio's licensors or channel partners have been advised of the possibility of such damages.<ltd3>}
\end{quote}

The \textbf{unilateral change} clause specifies the conditions under which the service provider could amend and modify the terms of service and/or the service itself. Such clause was always considered as potentially unfair. This is because the ECJ has not yet issued a judgment in this regard, though the Annex to the Directive contains several examples supporting such a qualification. Consider the following examples from the Twitter terms of service:
\begin{quote}
\sloppy \footnotesize
\texttt{<ch2>As such, the Services may change from time to time, at our discretion.</ch2>}
\end{quote}
\begin{quote}
\sloppy \footnotesize
\texttt{<ch2>We also retain the right to create limits on use and storage at our sole discretion at any time.</ch2>}
\end{quote}
\begin{quote}
\sloppy \footnotesize
\texttt{<ch2>We may revise these Terms from time to time. The changes will not be retroactive, and the most current version of the Terms, which will always be at twitter.com/tos, will govern our relationship with you.</ch2>}
\end{quote}

The \textbf{unilateral termination} clause gives provider the right to suspend and/or terminate the service and/or the contract, and sometimes details the circumstances under which the provider claims to have a right to do so.
Unilateral termination clauses that specify reasons for termination were marked as potentially unfair. whereas clauses stipulating that the service provider may suspend or terminate the service at any time for any or no reasons and/or without notice were marked as clearly unfair. That is the case in the three following examples, taken from the Dropbox and Academia terms of use, respectively:
\begin{quote}
\sloppy \footnotesize
\texttt{<ter2> We reserve the right to suspend or terminate your access to the Services with notice to you if:
(a) you're in breach of these Terms,
(b) you're using the Services in a manner that would cause a real risk of harm or loss to us or other users, or
(c) you don't have a Paid Account and haven't accessed our Services for 12 consecutive months.</ter2>}
\end{quote}
\begin{quote}
\sloppy \footnotesize
\texttt{<ter3>Academia.edu reserves the right, at its sole discretion, to discontinue or terminate the Site and Services and to terminate these Terms, at any time and without prior notice.</ter3>}
\end{quote}

The \textbf{contract by using} clause stipulates that the consumer is bound by the terms of use of a specific service, simply by using the service, without even being required to mark that he or she has read and accepted them.
We always marked such clauses  as potentially unfair. The reason for this choice is that a good argument can be offered for  these clauses to be unfair,  because they originate an imbalance in rights and duties of the parties, but  this argument  has no decisive authoritative backing yet, since  the ECJ has never assessed a clause of this type.
 Consider an example taken from the Spotify terms and conditions of use:
\begin{quote}
\sloppy \footnotesize
\texttt{<use2>By signing up or otherwise using the Spotify service, websites, and software applications (together, the ``Spotify Service'' or ``Service''), or accessing any content or material that is made available by Spotify through the Service (the ``Content'') you are entering into a binding contract with the Spotify entity indicated at the bottom of this document.</use2>}
\end{quote}

The  \textbf{content removal}  gives the provider a right to modify/delete user's content, including in-app purchases, and sometimes specifies the conditions under which the service provider may do so.
As in the case of unilateral termination, clauses that indicate conditions for content removal were marked as potentially unfair, whereas clauses stipulating that the service provider may remove content in his full discretion, and/or at any time for any or no reasons and/or without notice nor possibility to retrieve the content were marked as clearly unfair. For instance, consider the following examples, taken from Facebook's and Spotify's terms of use:
\begin{quote}
\sloppy \footnotesize
\texttt{<cr2> If you select a username or similar identifier for your account or Page, we reserve the right to remove or reclaim it if we believe it is appropriate (such as when a trademark owner complains about a username that does not closely relate to a user's actual name).</cr2>}
\end{quote}
\begin{quote}
\sloppy \footnotesize
\texttt{<cr2> We can remove any content or information you post on Facebook if we believe that it violates this Statement or our policies.</cr2>}
\end{quote}
\begin{quote}
\sloppy \footnotesize
\texttt{<cr3>In all cases, Spotify reserves the right to remove or disable access to any User Content for any or no reason, including but not limited to, User Content that, in Spotify's sole discretion, violates the Agreements. Spotify may take these actions without prior notification to you or any third party.</cr3>}
\end{quote}

The \textbf{arbitration} clause  requires or allows the parties to resolve their disputes through an arbitration process, before the case could go to court. It is therefore considered a kind of forum selection clause. However, such a clause may or may not specify that arbitration should occur within a specific jurisdiction.
Clauses stipulating that the arbitration should  (1) take place in a state other then the state of consumer's residence and/or (2) be based not on law but on arbiter's discretion were marked as clearly unfair.
As an illustration, consider the following clause of the Rovio terms of use:
\begin{quote}
\sloppy \footnotesize
\texttt{<j1> <a3>Any dispute, controversy or claim arising out of or relating to this EULA or the breach, termination or validity thereof shall be finally settled at Rovio's discretion (i) at your domicile's competent courts; or (ii) by arbitration in accordance with the Rules for Expedited Arbitration of the Arbitration Institute of the Finland Chamber of Commerce. The arbitration shall be conducted in Helsinki, Finland, in the English language.</a3> </j1>}
\end{quote}
Notice that the above clause concerns both jurisdiction and arbitration (notice the use of nested tags).
%
%
Clauses defining arbitration as fully optional would have to be marked as clearly fair. However, our corpus does not contain any example of fully optional arbitration clause.
Thus,  all arbitration clauses were marked as potentially unfair. An example is the following segment of Amazon's terms of service:
\begin{quote}
\sloppy \footnotesize
\texttt{<a2>Any dispute or claim relating in any way to your use of any Amazon Service, or to any products or services sold or distributed by Amazon or through Amazon.com will be resolved by binding arbitration, rather than in court, except that you may assert claims in small claims court if your claims qualify. The Federal Arbitration Act and federal arbitration law apply to this agreement.</a2>}
\end{quote}

\subsection{Corpus statistics}

The final corpus contains 12,011 sentences\footnote{The segmentation into sentences was obtained with Stanford CoreNLP suite.} overall, 1,032 of which (8.6\%) were labeled as positive, thus containing a potentially unfair clause. The distribution of the different categories across the 50 documents is reported in Table~\ref{tab:corpus_statistics}: arbitration clauses are the least common, being present in 28 documents only, whereas all the other categories appear in at least 40 out of 50 documents. Limitation of liability and unilateral termination categories represent more than half of the total potentially unfair clauses. The percentage of potentially unfair clauses in each document is quite heterogeneous, ranging from 3.3\% (Microsoft) up to 16.2\% (TrueCaller).

\begin{table}[!bt]
\begin{center}
\caption{Corpus statistics. For each category of clause unfaireness, we report the overal number of clauses and the number of documents they appear in.\label{tab:corpus_statistics}}
\begin{tabular}{|c|c|c|}
    \hline
    Type of clause & \# clauses & \# documents \\
    \hline
    Arbitration   			     & 44  & 28 \\
    Unilateral change  		   & 188 & 49	\\
    Content removal  		     & 118 & 45	\\
    Jurisdiction			       & 68  & 40	\\
    Choice of law 			     & 70  & 47	\\
    Limitation of liability  & 296 & 49	\\
    Unilateral termination   & 236 & 48	\\
    Contract by using  		   & 117 & 48	\\
    \hline
  \end{tabular}
\end{center}
\end{table}

\section{Machine Learning Methodology} \label{sec:methodology}
In this section we briefly describe the representation and learning methods we used in our study. We address two different tasks: a detection task, aimed at predicting whether a given sentence contains a (potentially) unfair clause, and a classification task, aimed at predicting the category an unfair clause belongs to, which indeed could be a valuable piece of information to a potential user.

\subsection{Learning algorithms}

We address the problem of detecting potentially unfair contract clauses as a sentence classification task. Such a task could be tackled by treating sentences independently of one another ({\it sentence-wide} classification). This is the most standard and classic approach in machine learning, traditionally addressed by methods such as Support Vector Machines or Artificial Neural Networks (including recent deep learning approaches).

Alternatively, one could take into account the structure of the document, in particular the \textit{sequence} of sentences, so as to perform a \textit{collective} classification. Because it is not uncommon for unfair clauses to span across consecutive sentences in a document, this approach could also have some advantages.

In sentence-wide classification the problem can be formalized as follows. Given a sentence, the goal is to classify it as \textit{positive} if it contains a potentially unfair clause, or \textit{negative} otherwise. Within this setting, a machine learning classifier is trained with a data set $\mathcal{D} = \{(x_i,y_i)\}_{i=1}^N$, which consists of a collection of $N$ pairs, where $x_i$ encodes some representation of a sentence, and $y_i$ is its corresponding (positive or negative) class. 

In collective classification, the data set consists of a collection of $M$ \textit{documents}, represented as sequences of sentences: $$\mathcal{D} = \{ d_j = \{(x_{1}^j,y_{1}^j), \ldots, (x_{k_j}^j,y_{k_j}^j)\} \}_{j=1}^M,$$ where the $j$-th document contains $k_j$ sentences. 

Different machine learning systems can be developed for each classification setup, according to the learning framework and to the features employed to represent each sentence. As for the learning methodology, for sentence-wide classification in this paper we compare Support Vector Machines (SVMs) \citep{Joachims1998} with some recent deep learning architectures, namely Convolutional Neural Networks (CNNs)~\citep{Kim2014} and Long-Short Term Memory Networks (LSTMs)~\citep{Graves2005}. For collective classification, we rely on structured Support Vector Machines, and in particular on SVM-HMMs, which combine SVMs with Hidden Markov Models~\citep{Tsochantaridis2005}, by jointly assigning a label to each element in a given sequence (in our case, to each sentence in the considered document).

\subsection{Sentence representation}

As for the features represented to encode sentences, in an effort to make our method as general as possible, we decided to opt for traditional features for text categorization, excluding other, possibly more sophisticated, handcrafted features.

One of the most classic, yet still widely used, set of features for text categorization, is the well-known \textit{bag-of-words} (BoW) model. In such a model, one feature is associated to each word in the vocabulary: the value of such feature is either zero, if the word does not appear in the sentence, or other than zero, if it does. Such a value is usually computed as the TF-IDF score, that is product of the number of occurrences of the word in the sentence (Term Frequency, TF) by a term that strengthen the contribution of infrequent words (Inverse Document Frequency, IDF)~\citep{Sebastiani2002}.

The BoW model can be extended to consider also $n$-grams, i.e., consecutive word combinations, rather than simple words, so that the order of the words in the sentences is (at least locally) exploited. Grammatical information can be included as well, by constructing a bag of part-of-speech tags, i.e., word categories such as nouns, verbs, etc.~\citep{Leopold2002}.
Despite their simplicity, BoW features are very informative, as they encode the lexical information of a sentence, and thus represent a challenging baseline in those cases where the presence of some keywords and phrases is highly discriminative for the categorization of sentences.

\begin{figure}
\begin{center}
\begin{tikzpicture}[level distance=0.7cm,scale=.7]
\Tree [.S [.NP [.NP [.NNS Portions ] ] [.PP [.IN of ] [.NP [.DT the ] [.NN Amazon ] [.NNS services ] ] ] ] [.VP [.VBP operate ] [.PP [.IN under ] [.NP [.NP [.NN license ] ] [.PP [.IN of ] [.NP [.QP [.CD one ] [.CC or ] [.JJR more ] [.NNS patents ] ] ] ] ] ] ] [.. . ] ]
\end{tikzpicture}
\caption{An example of a constituency parse tree for a sentence in our corpus.\label{fig:tree}}
\end{center}
\end{figure}
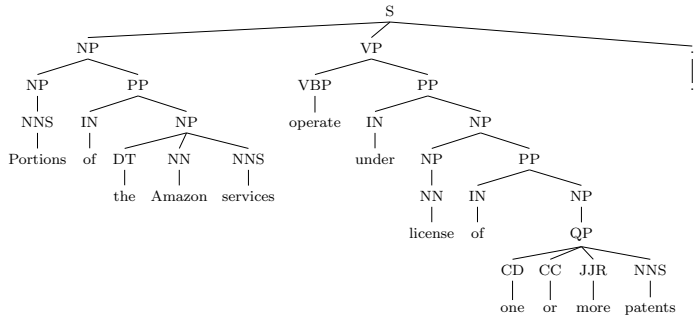

A second approach we consider for the representation of a sentence is to exploit a constituency parse tree, which naturally encodes the \textit{structure} of the sentence (see Figure~\ref{fig:tree}) by describing the grammatical relations between sentence portions through a tree. Similarity between tree structures can be exploited with \textit{tree kernels}~\citep{Moschitti2006} (TK).
A TK  consists of a \textit{similarity measure} between two trees, which takes into account the number of common substructures or \textit{fragments}. Different definitions of fragments induce different TK functions. In our study we use the SubSet Tree Kernel (SSTK)~\citep{Collins2002} which counts as fragments those subtrees of the constituency parse tree terminating either at the leaves or at the level of non-terminal symbols. SSTK have been shown to outperform other TK functions in several argumentation mining sub-tasks~\citep{MARGOT}.

A third approach for sentence representation is based on \textit{word embeddings}~\citep{Mikolov2013}, a popular technique that has been recently developed in the context of neural language models and deep learning applications.
Neural networks such as CNNs and LSTMs can handle textual input, by converting it into a sequence of identifiers (one for each different word): it is the neural network, then, which directly learns a vector representation (named embedding) of words and sentences.

\section{Experimental Results} \label{sec:results}
We evaluated and compared several different machine learning systems on the data set presented in Section~\ref{sec:dataset}. Each document was segmented into sentences, tokenized and parsed with the Stanford CoreNLP tool.\footnote{\url{https://stanfordnlp.github.io/CoreNLP/}}
Sentences and text fragments shorter than 5 words were discarded.
We obtained a total of 9,414 sentences, 1,032 of which (11.0\%) were labeled as positive, thus containing a potentially unfair clause.
We run experiments following the \textit{leave-one-document-out} (LOO) procedure, in which each document in the corpus, in turn, is used as test set, leaving the remaining documents for training set (4/5) and validation set (1/5) for model selection. To quantitatively evaluate the different systems, we computed precision ($P$) as the fraction of positive predictions, which are actually labeled as positive, recall ($R$) as the fraction of positive examples that are correctly detected, and finally $F_1$ as the harmonic mean between precision and recall ($F_1 = \frac{2PR}{P+R}$). These performance measurements were aggregated using the macro-average over documents~\citep{Sebastiani2002}.

For the first task (potentially unfair clause detection) we compared several systems. The problem is formulated as a binary classification task, where the positive class is either the union of all potentially unfair sentences, or the set of potentially unfair clauses of a single category, as described below. We considered the following systems:
\begin{enumerate}[C1:]
  \item a single SVM exploiting BoW (unigrams and bigrams for words and part-of-speech tags);
  \item a combination of eight SVMs (same features as above), each considering a single unfairness category as the positive class, whereby a sentence is predicted as potentially unfair if at least one of the SVMs predicts it as such;
  \item a single SVM exploiting TK for sentence representation;
  \item a CNN trained from plain word sequences;
  \item an LSTM trained from plain word sequences;
  \item an SVM-HMM performing collective classification of sentences in a document (same features as C1);
  \item a combination of eight SVM-HMMs, each performing collective classification of sentences in a document on a single unfairness category as the positive class (same setting as C2);
  \item an ensemble method, that combines the output of C1, C2, C3, C6 and C7 with a voting procedure (sentence predictive as positive if at least 3 systems out of 5 classify it as such).
\end{enumerate}
As a reference for the complexity of the task, we also report the performance of the following baselines: a \textit{random} classifier, which predicts potentially unfair clauses at random,\footnote{Sampling takes into account the class distribution in the training set.}
and an \textit{always positive} baseline, which classifies every sentence as potentially unfair. For all the classifiers, the validation set was used to select the best hyper-parameters. For all SVMs we used a linear kernel, thus optimizing the $C$ parameter only. For SVM-HMM we used an order of dependencies equal to 2 and 1 for transitions and emissions, respectively; different from SVMs, we also used trigrams besides unigrams and bigrams, as they slightly increased performance. For CNNs, we considered one layer with 64 filters of size equal to 3, followed by two fully connected layers with 32 and 16 neurons, respectively. We applied dropout equal to 0.5, batch size equal to 16. An embedding of size 64 was learned after the input layer. For LSTMs, we considered a 2-layer network with 64 and 32 cells, respectively, with 0.25 dropout and mini-batch size equal to 16. An embedding of size 32 was learned after the input layer. Both for CNNs and LSTMs, no improvement was observed if using pre-trained word embeddings.

Table~\ref{tab:results_comparison} shows the results achieved by each of these variants.
If we exclude the ensemble approach, the best classifier in terms of $F_1$ results to be C2, that is the system combining one different SVM trained for each unfairness category, with a precision above 80\%, and a recall of 78\%. The structured SVMs exploiting the sequentiality of the sentences achieve slightly lower results, yet very interestingly the results of the sentence-wise and document-wise approaches are different across different documents.
Moreover, the worse performance associated with TK suggests that the syntactic structure of the sentence is less informative than the lexical information captured by $n$-grams. This makes the task of detecting unfair clauses different from other text retrieval problems in the legal domain, such as, for example, the detection of claims and arguments~\citep{TOIT2016}. As for CNNs and LSTMs, the slightly worse performance with respect to the other approaches could also be ascribed to the limited size of the training set.

All these observations led us to the implementation of an ensemble method (C8), combining the five best performing approaches. This system achieves an $F_1$ of around 81\%, thus beating all the competitors. Such a result is particularly interesting, because it confirms that the different systems capture complementary information for the detection of potentially unfair clauses. The ensemble method correctly detects over 75\% of the potentially unfair clauses of all the categories, from 76.6\% of Unilateral Change up to 89.7\% for Jurisdiction.

In order to gain a better understanding of which are the $n$-grams that contribute the most to the discrimination between fair and potentially unfair clauses, we computed the frequencies of 2-grams in both positive and negative support vectors of classifier C2, and we looked for those with the largest discrepancy in appearing in the positive class rather than in the negative one. These were some of the most significant 2-grams, according to such ranking: \textit{for any}, \textit{the right}, \textit{these terms}, \textit{any time}, \textit{at any}, \textit{right to}, \textit{reserves the}, \textit{we may}, \textit{liable for}, \textit{terminate your}, \textit{sole discretion}, \textit{the services}. This analysis confirms that the discriminative lexicon is quite general and widespread both across the different unfairness categories and the different types of services we considered.

\begin{table}
  \begin{center}
  \caption{Results on leave-one-document-out procedure. \label{tab:results_comparison}}
  \begin{tabular}{|c|c|c|c|c|}
    \hline
    Classifier & Method & $P$ & $R$ & $F_1$ \\
    \hline
    C1 & SVM -- Single Model       			& 0.729 & 0.830 & 0.769 \\
    C2 & SVM -- Combined Model     			& 0.806 & 0.779 & 0.784 \\
    C3 & Tree Kernels              			& 0.777 & 0.718 & 0.739 \\
    C4 & Convolutional Neural Networks		& 0.729 & 0.739 & 0.722 \\
    C5 & Long Short-Term Memory Networks	& 0.696 & 0.723 & 0.698 \\
    C6 & SVM-HMM -- Single Model			& 0.759 & 0.778 & 0.758 \\
    C7 & SVM-HMM -- Combined Model			& 0.848 & 0.720 & 0.772 \\
    C8 & Ensemble (C1+C2+C3+C6+C7)			& 0.828 & 0.798 & 0.806 \\
    \hline
    & Random Baseline           			& 0.125 & 0.125 & 0.125 \\
    & Always Positive Baseline  			& 0.123 & 1.000 & 0.217 \\
    \hline
  \end{tabular}
  \end{center}
\end{table}

The second task we considered is unfairness categorization, for which we employed eight SVM classifiers, each trained to discriminate between potentially unfair clauses of one category with respect to all the other categories. Note that this task differs from that addressed by the previously introduced classifiers, since in this case the classifiers is trained on potentially unfair clauses only. In Table~\ref{tab:tag_details} we report the precision, recall, and F$_1$ of such classifiers, one for each separate tag category, micro-averaged on the whole dataset. The results show that discriminating amongst the different categories is a simpler task, since the $F_1$ is larger than 74\% for all tags, and larger than 93\% four tags (jurisdiction, choice of law, limitation of liability, contract by using).

\begin{table}
  \begin{center}
  \caption{Micro-averaged precision, recall and F$_1$ of abusive clauses for each tag category.\label{tab:tag_details}}
  \begin{tabular}{|c|P{1.5cm}|P{1.5cm}|P{1.5cm}|}
    \hline
    Tag & Precision & Recall & F$_1$ \\
    \hline
    Arbitration             & 0.832 & 0.814 & 0.823 \\
    Unilateral change       & 0.832 & 0.814 & 0.823 \\
    Content removal         & 0.713 & 0.780 & 0.745 \\
    Jurisdiction            & 1.000 & 0.941 & 0.970 \\
    Choice of law           & 0.984 & 0.886 & 0.932 \\
    Limitation of liability & 0.961 & 0.905 & 0.932 \\
    Unilateral termination  & 0.786 & 0.932 & 0.853 \\
    Contract by using       & 0.949 & 0.957 & 0.953 \\
    \hline
  \end{tabular}
  \end{center}
\end{table}

\section{The CLAUDETTE Web Server} \label{sec:server}
The proposed approach was implemented and developed within a web server, reachable at the address \url{http://155.185.228.137/claudette/}, so as to produce a prototype system that users can easily access and test.

\begin{figure}[!t]
\begin{center}
\includegraphics[width=0.7\textwidth]{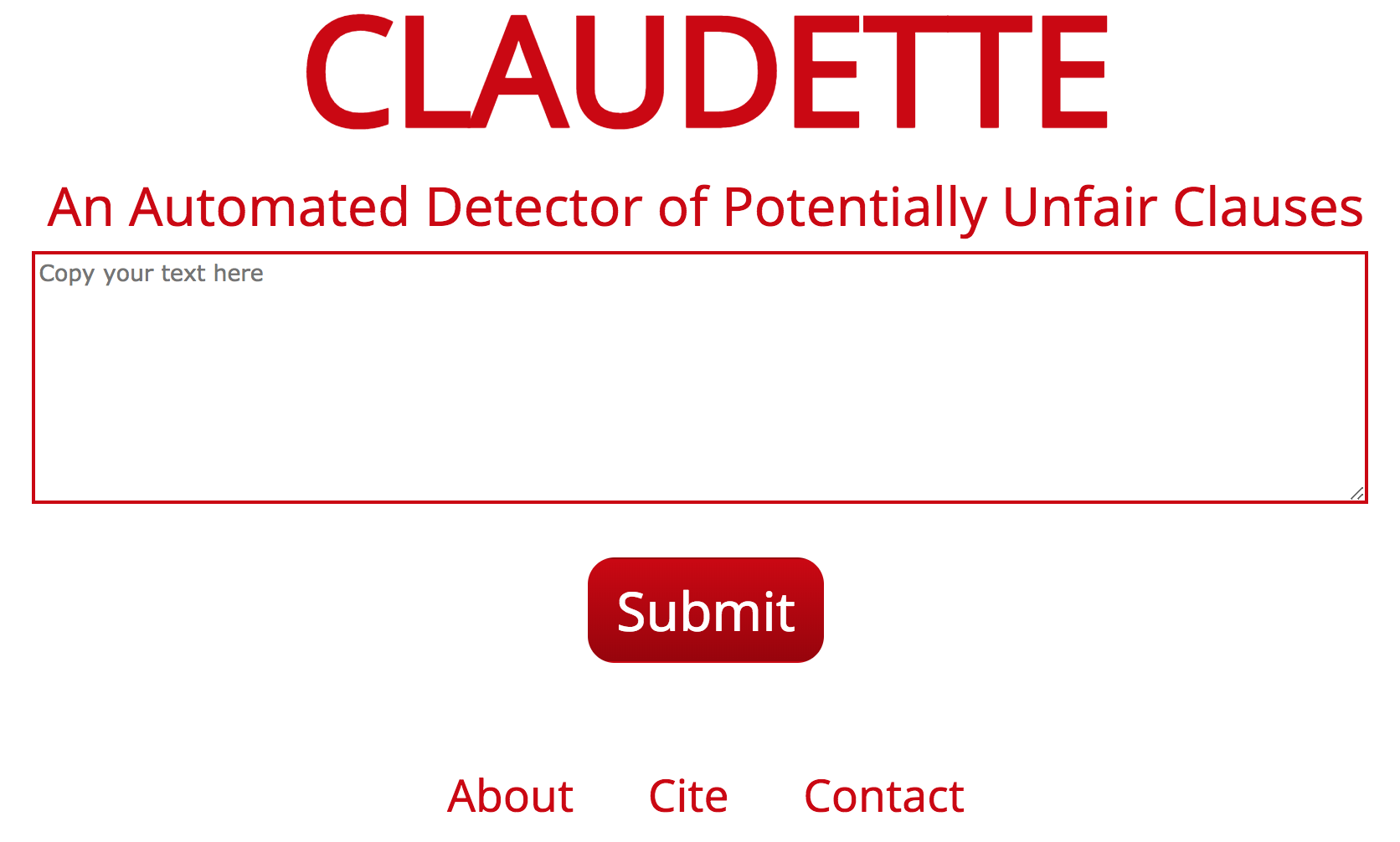}
\caption{The interface of the CLAUDETTE web server, consisting of a box where a user can copy-paste the text of a terms of service.\label{fig:claudette}}
\end{center}
\end{figure}

\begin{figure}[!t]
\begin{center}
\includegraphics[width=\textwidth]{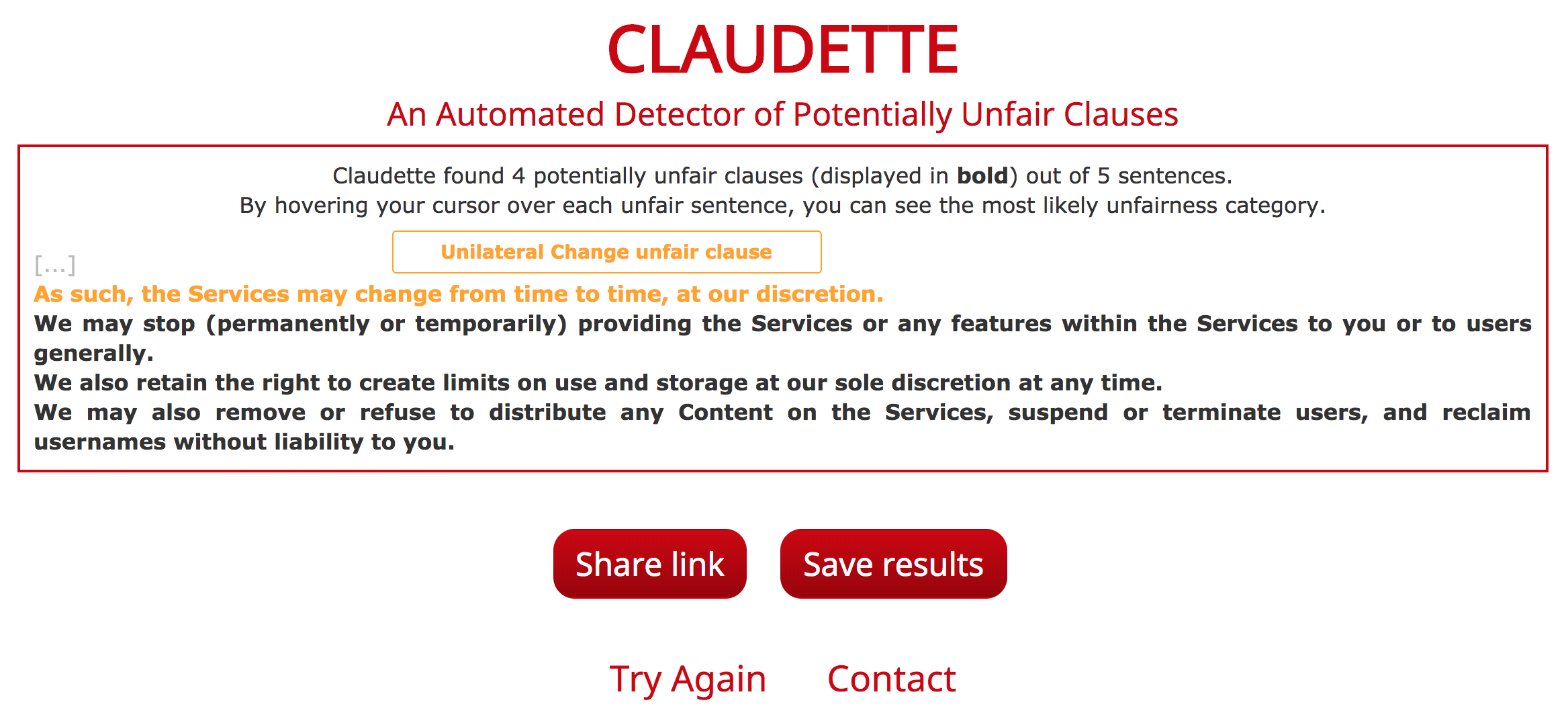}
\caption{Results of a query to the CLAUDETTE web server. Hovering over a detected clause with the pointer provides an indication of the type of potentially unfair clause. In this example the detected clauses are predicted to be of types \textit{unilateral change}, \textit{unilateral termination}, and \textit{content removal}, and the cursor was left hovering over the first potentially unfair clause.\label{fig:claudette-output}}
\end{center}
\end{figure}

As shown in Figure~\ref{fig:claudette}, the interface is easy to use. %
A user only needs to paste the text to be analyzed and push a button. The system will then produce an output file that highlights the sentences predicted to contain a potentially unfair clause. The output will also indicate the predicated category the unfair clause belongs to, as illustrated in Figure~\ref{fig:claudette-output}. The output of the system can be obtained in several formats including HTML, XML, JSON, and plain text.

For this online service, for the detection stage we implemented only one system (namely, classifier C2) rather than the ensemble method, because it resulted to be a much more efficient solution, despite producing a slightly lower performance accuracy.

\section{Related Work} \label{sec:related}
The use of artificial intelligence, machine learning and natural language processing techniques in the analysis and classification of legal documents is gaining a growing interest~\citep{ashley2017artificial}. Among others,~\cite{Moens2007} proposed a pipeline of steps for the extraction of arguments from legal documents, exploiting supervised classifiers and context-free grammars, whereas~\cite{Biagioli2005} proposed to employ multi-class SVM for the identification of significant text portions in normative texts. Recent approaches have focused on the detection of claims~\citep{Lippi2015aicol} and of cited facts and principles in legal judgments~\citep{Shulayeva2017}, as well as on the prediction of judicial decisions~\citep{Aletras2016}. A case study regarding the construction of legal arguments in the legal determinations of vaccine/injury compensation compliance using natural language tools was given by~\cite{Ashley2013}. 
Finally, privacy policies represent another strictly related application where machine learning approaches have proved  effective, as discussed by~\cite{Fabian2017} and references therein, as well as ~\cite{harkous2018polisis}.
Besides the work by \cite{Lippi2017JURIX} which we discussed in the introduction, we are not aware of other recent  text processing applications in consumer law.

\section{Conclusions} \label{sec:conclusion}
Our study investigates the use of machine learning and natural language methods for the automated detection of potentially unfair clauses in online contracts. We addressed two tasks: clause detection and clause type classification. For clause detection, our results are very encouraging: using a relatively small training set we could automatically detect over 80\% clauses, with an 80\% precision. The categorization task turned out to be simpler. Given that most unfair clauses are currently hidden within long and hardly readable ToS, the recall and precision offered by our approach may  already be significant enough to enable useful applications. 

It is interesting to notice the comparatively better performance of the BoW approach with respect to other more sophisticated approaches. That is in agreement with the surveyed literature, where classic lexical approaches such as BoW still represent a crucial ingredient of automated systems. It is also worth remarking that an ensemble method produced the best performance, thus indicating that different machine learning approaches are capable of capturing different characteristics of potentially unfair clauses.


This study was motivated by a long-term goal such as the pursuit of effective consumer protection by way of tools that support consumers and their organizations in detecting unfair contractual clauses. We plan to extend our analysis to other machine learning methods that could contribute to such tools. 
In particular, we are studying ways to exploit contextual information, since it was pointed out that the fairness of clauses might very well depend on the context. For example, a potentially unfair jurisdiction clause might actually be fair according to EU regulation if is followed by a paragraph stipulating relevant exceptions according to the user's country of residence.

As a further, challenging line of research, we are planning to apply similar methodologies also to privacy policies: an important area of consumer protection that has recently gained media focus due to its enormous implications not only for individuals but also for society at large.

\bibliographystyle{spbasic}      
\bibliography{jurix}  

\end{document}